\documentclass{article}
\usepackage{spconf,amsmath,graphicx}
\usepackage[printonlyused,withpage]{acronym}
\usepackage{amssymb}
\usepackage{amsmath}
\usepackage{algorithm}
\usepackage{booktabs}
\usepackage{multirow}
\usepackage[noend]{algpseudocode}

\usepackage{caption}
\usepackage{subcaption}
\usepackage{graphicx}
\usepackage{soul}
\usepackage[dvipsnames]{xcolor}

\usepackage{fancybox}

\title{
Towards Interpretability of Speech Pause in Dementia Detection using Adversarial Learning
}
\name{Youxiang Zhu$^1$, Bang Tran$^1$, Xiaohui Liang$^1$, John A. Batsis$^2$, and Robert M. Roth$^3$\thanks{This research is funded by the US National Institutes of Health National Institute on Aging, under grant No. R01AG067416.}}
\address{
  $^1$Department of Computer Science, University of Massachusetts Boston, MA, USA\\
  $^{2}$ School of Medicine, University of North Carolina, Chapel Hill, NC, USA\\
  $^{3}$ Geisel School of Medicine at Dartmouth, Lebanon, NH, USA
}

\begin{document}
%\ninept
%
\maketitle

% \onecolumn
%
\begin{abstract}
Speech pause is an effective biomarker in dementia detection. Recent deep learning models have exploited speech pauses to achieve highly accurate dementia detection, but have not exploited the interpretability of speech pauses, i.e., what and how positions and lengths of speech pauses affect the result of dementia detection. In this paper, we will study the positions and lengths of dementia-sensitive pauses using adversarial learning approaches. Specifically, we first utilize an adversarial attack approach by adding the perturbation to the speech pauses of the testing samples, aiming to reduce the confidence levels of the detection model. Then, we apply an adversarial training approach to evaluate the impact of the perturbation in training samples on the detection model. We examine the interpretability from the perspectives of model accuracy, pause context, and pause length. We found that some pauses are more sensitive to dementia than other pauses from the model's perspective, e.g., speech pauses near to the verb ``is". Increasing lengths of sensitive pauses or adding sensitive pauses leads the model inference to \ac{AD}, while decreasing the lengths of sensitive pauses or deleting sensitive pauses leads to non-AD.

\end{abstract}
\begin{keywords}
Speech pauses, interpretability, acoustic feature, dementia detection, spontaneous speech
\end{keywords}
\section{Introduction}
\label{sec:intro}

\begin{figure*}[ht!]
    \centering
    \includegraphics[width=0.8\textwidth]{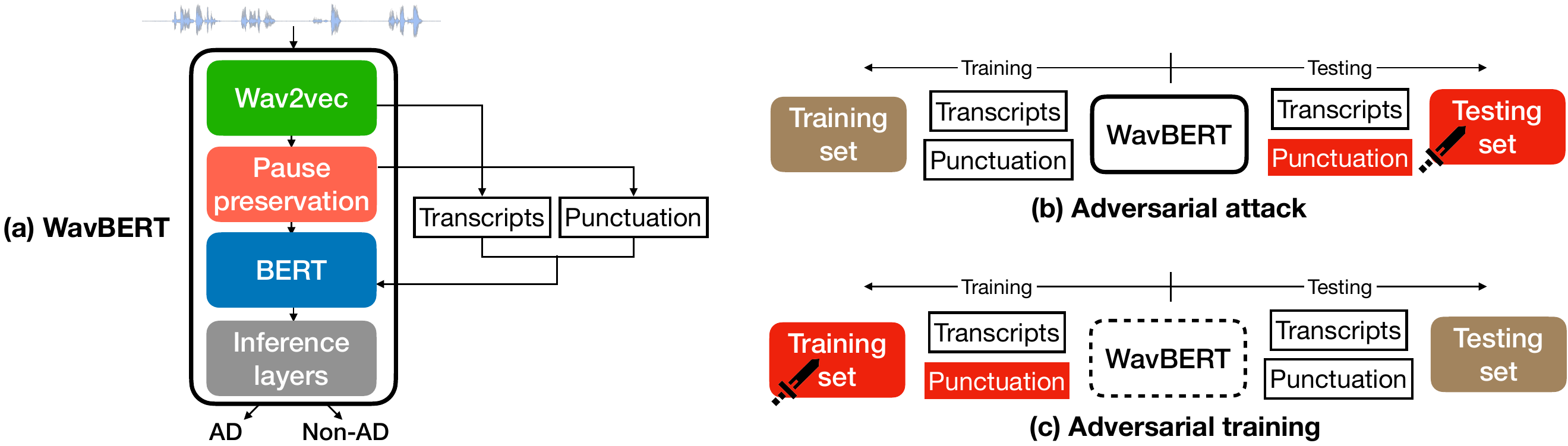}
    \caption{The description of WavBERT model (a), the adversarial attack approach (b), and the adversarial training approach (c).}
    \label{fig:model}
    \vspace{-0.3cm}
\end{figure*}

Exploiting spontaneous speech for dementia detection has shown promise in recent years. In the \ac{ADReSS} challenge~\cite{luz2020alzheimer,luz2021detecting}, spontaneous speech datasets were collected using a Cookie Theft picture description task~\cite{goodglass2001bdae}, and researchers have developed deep learning models
for achieving a promising classification accuracy in differentiating \ac{AD} from non-AD patients~\cite{yuan2020disfluencies,yuan2020pauses}. One common acoustic biomarker these successful models have exploited is the speech pauses~\cite{yuan2020disfluencies, yuan2020pauses, sluis2020automated}, which have been widely studied in the medical research~\cite{pastoriza2021speech, davis2009examining}. However, these deep learning models have not analyzed the impact of the positions and lengths of the speech pauses on the detection of dementia. Thus, it remains unknown that to what extent these models understand the speech pauses to make the decision.

Recent deep transfer learning studies have significantly advanced our understanding of the domains of image, language, and speech. Deep transfer learning models learn features automatically from large-scale datasets, and these auto-learned features can be more representative than conventional handcrafted features. In dementia detection, such models produced a high accuracy~\cite{balagopalan2020bert,yuan2020disfluencies,yuan2020pauses}, but they often lacked interpretability, limiting the trustworthiness of their medical applications. Attention and visualization mechanisms are common methods to enhance the interpretability of deep learning models~\cite{ribeiro2016should,goyal2019counterfactual,simonyan2013deep,clark2019does, liu2019linguistic, che2015distilling}. However, these methods do not apply to the speech-based dementia detection scenario because of the complex relationship between input data and labels. In the image domain, labels are directly related to the input data, e.g., a dog and a dog image. Humans can evaluate whether the interpretability results are consistent with human knowledge, e.g., the model assigns large weights to the dog's body area to recognize the dog. In contrast, dementia labels are derived from a standard and complex assessment, which is not directly related to the speech data. Humans hardly evaluate the interpretability results on dementia-related speech. Therefore, we evaluated the interpretability of speech pauses, rather than words expressed, for dementia detection.

The WavBERT dementia detection model~\cite{zhu2021wavbert} is the first model preserving the pause in the text without using the manual transcription. WavBERT uses Wav2vec~\cite{baevski2020wav2vec} to measure the lengths of speech pauses and represents them with punctuation marks. In this paper, we use adversarial learning and WavBERT to study the speech pauses. Specifically, we first propose an adversarial attack approach where we generate adversarial samples from original testing samples by adding, replacing, and deleting the punctuation marks. We test the WavBERT model using the adversarial samples and identify the perturbation action that leads to the most significant change in the model's confidence. In an adversarial training approach, other than training with the original samples, we additionally train the WavBERT model with two groups of adversarial samples, one includes samples positively influencing the confidence, and the other includes samples negatively influencing the confidence. We then investigate how the adversarial training affects the model on the testing samples. The contributions of this paper are three-fold.

First, we exploit adversarial learning to study the interpretability of speech pauses in dementia detection. We propose methods to evaluate the impact of speech pauses from the perspectives of model accuracy, pause context and length.

Second, we found that the speech pauses between ``is" and another verb is dementia-sensitive, and confirmed this observation with the training set. We also found that adding pauses at sensitive places or increasing the lengths of sensitive pauses leads the model inference toward AD, while deleting sensitive pauses or decreasing the lengths of sensitive pauses leads the model inference toward non-AD.

Third, we found that the misspelling errors from Wav2vec introduced uncommon tokens, leading to misinterpretation of the speech pauses. Such problem is inevitable if using a speech recognition model and a language model, and might be resolved with large-scale speech representation.

\section{WavBERT - Pause preservation}

WavBERT~\cite{zhu2021wavbert} utilizes speech pauses and transcripts for dementia detection. It transcribes the speech audio to transcript using Wav2vec, and this transcript does not have punctuation marks. WavBERT uses the number of blank tokens from the intermediate results of Wav2vec to define two levels of pauses: longer sentence-level pause and shorter in-sentence pause. A ``period" mark is inserted into the transcript if a sentence-level pause is determined; a ``comma" mark is inserted into the transcript if an in-sentence pause is determined. After adding the punctuation ``period" and ``comma" into the transcripts, WavBERT feeds the transcripts to BERT for \ac{AD} classification. More details can be found in Figure~\ref{fig:model} and the paper~\cite{zhu2021wavbert}. 
In this paper, we add one more level of in-sentence pause and represent it with ``semicolon." The length of such pause is longer than ``comma" and shorter than ``period."

\begin{figure*}[h!]
    \centering
    \includegraphics[width=0.87\textwidth]{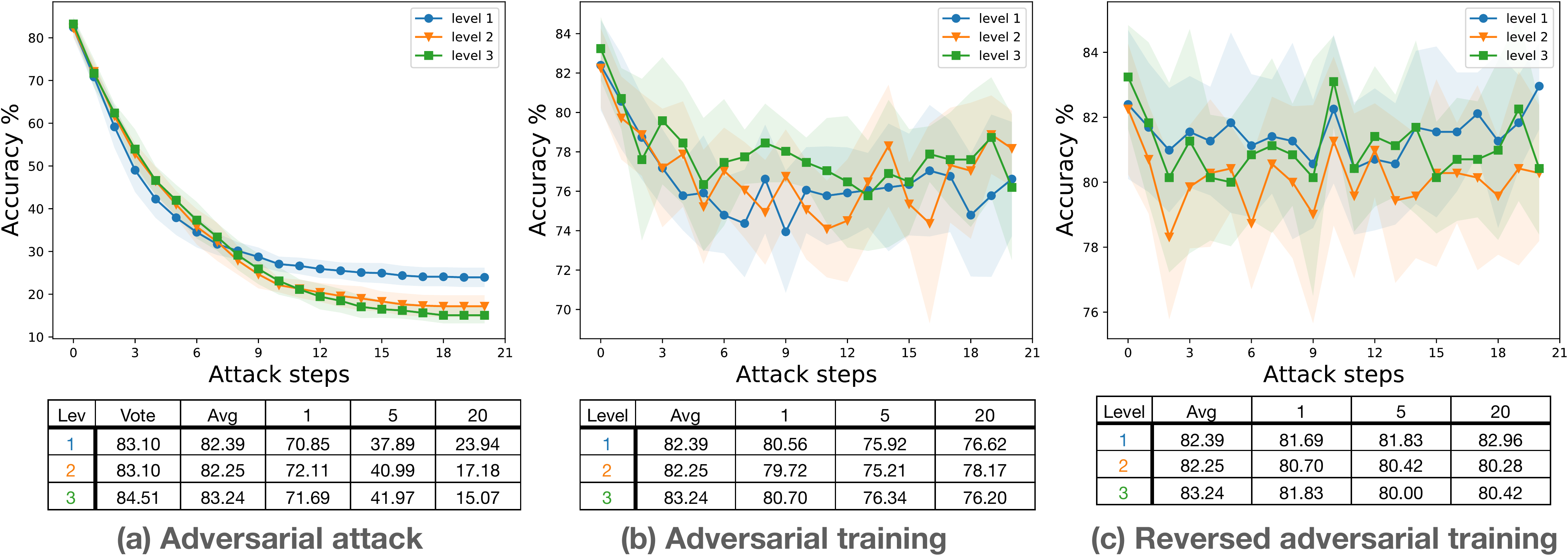}
    \caption{Impact of adversarial attack/training.
    Level 1: period. Level 2:  comma + period. Level 3: comma + semi-colon + period}
    \label{fig:result}
    \vspace{-0.5cm}
\end{figure*}

\section{Proposed adversarial learning}

We present two adversarial learning approaches to exploit the impact of speech pauses on dementia detection.

\textbf{Adversarial attack.} An adversarial attack adds perturbation to the punctuation marks. Note that, other adversarial attacks on BERT considered the entire vocabulary as the attack space~\cite{garg2020bae, li2020bert}. In our case, the attack space only includes punctuation marks, and the perturbation actions are adding, deleting, and replacing. As shown in Figure~\ref{fig:model}(b), we adopt the same training phase of the original WavBERT. To launch the attack, we choose an original testing sample $s_{0}$, and generate a set of adversarial samples by using a single action of adding, deleting, or replacing. For adding, we add one punctuation mark between two neighboring words where no punctuation mark exists; for deleting, we delete an existing punctuation mark; and for replacing, we replace an existing punctuation mark with a different punctuation mark, e.g., replacing ``comma" with ``period." We define 
a \emph{\textbf{confidence level}} as $l_{AD} = o_{AD}(s)$ for an AD sample $s$, and $l_{non-AD} = o_{non-AD}(s)$ for a non-AD sample $s$, where $o_{AD}(s)$ and $o_{non-AD}(s)$ are the logit output by the model for \ac{AD} and non-AD labels. Among the samples generated from $s_0$, we identify a sample $s_1$ as the most effective sample if the confidence level of $s_{1}$ is the lowest. We call $s_1$ as the step-1 sample, and then iteratively launch the attack on $s_{1}$ to obtain the step-2 sample $s_{2}$. The attack will be launched for a maximum of 20 steps unless the confidence level cannot be lowered. The details are shown in Algorithm~\ref{alg:1}.

\begin{algorithm}[h!]
    \caption{Adversarial attack $s_i\rightarrow s_{i+1}$}
    \label{alg:1}
    \hspace*{\algorithmicindent} \textbf{Input:}  A step-$i$ sample $s_i$ and a model $M$\\
    \hspace*{\algorithmicindent} \textbf{Output:} A step-$(i+1)$ sample $s_{i+1}$
    \begin{algorithmic}[1]
    
    \State{$s_{i+1} = \mbox{NULL}$, confidence level $l_i = M(s_i)$}
    
    \State{Generate a set of adversarial samples from $s_i$ using one perturbation action: $C_{i+1} = \{s_{i+1,1},s_{i+1,2}, ..., s_{i+1,n}\}$.}
    
    \For{each sample $s_{i+1,j}$ in $C_{i+1}$}
         \State{$l_{i+1,j} = M(s_{i+1,j})$. }
         \If{$l_{i+1,j} < l_i$ }
            \State{$l_i=l_{i+1,j}$, $s_{i+1} := s_{i+1,j}$}
         \EndIf
    \EndFor
    
    \State{Return $s_{i+1}$}
    
    \end{algorithmic}
\end{algorithm}

\textbf{Adversarial training.} In another adversarial learning approach, we generate adversarial samples and apply adversarial training to the WavBERT, as shown in Figure~\ref{fig:model}(c). Specifically, we add perturbations to the original training samples of WavBERT. The perturbation actions are the same as the previous adversarial attack approach. In each attack step on a sample, we choose the adversarial sample that has the \textbf{lowest} confidence level. The attack is iteratively launched for a maximum of 20 steps unless the confidence level cannot be lowered. Finally, the original training samples and perturbed samples are used to train the WavBERT. The updated model is expected to be less effective because adversarial samples produce a negative impact in the training phase. We further propose a reversed adversarial training approach. We generate an adversarial sample from an original sample that has the \textbf{highest} confidence level. In this way, we envision that the reversed adversarial training makes the model more effective.

\textbf{Interpretability.} We study the interpretability of speech pauses from the following three perspectives.

\emph{Model accuracy.} In the adversarial attack approach, we can evaluate how much the perturbation affects the model accuracy. In the adversarial training approach, the changes of model accuracy reveal how well the model understands the speech pauses in the original samples.

\emph{Pause context.} In each step of the adversarial attack, we choose the perturbation that causes the most negative impact on the confidence level. We can analyze the statistics of the neighboring words of all the chosen punctuation marks, and identify the dementia-sensitive pauses based on the context.

\emph{Pause length.} We divide the perturbation actions into two groups. Group 1 involves adding and replacing a short pause with a long pause, and Group 2 involves deleting and replacing a long pause with a short pause. Group 1 shifts the result towards AD, and Group 2 shifts the result towards non-AD.

\begin{table*}[!htbp]
\centering
\caption{Context analysis on frequency of pauses. `\#' represents a pause. $r$ represents the range, and $s$ represents the step}
%\vspace{-0.2cm}
\label{tab:word}
\resizebox{\textwidth}{!}{%
\begin{tabular}{ccccccc|cccccc}
\hline
%  & \multicolumn{6}{c|}{$r=1$} & \multicolumn{6}{c}{$r=2$} \\
 & \multicolumn{2}{c}{$r=1$, $s=1$} & \multicolumn{2}{c}{$r=1$, $s=5$} & \multicolumn{2}{c|}{$r=1$, $s=20$} & \multicolumn{2}{c|}{$r=2$, $s=1$} & \multicolumn{2}{c}{$r=2$, $s=5$} & \multicolumn{2}{c}{$r=2$, $s=20$} \\ \hline
\multirow{5}{*}{level 1} & it \# looks & 19 & of \# the & 40 & the \# sink & 102 & the mother \# and the & 10 & \underline{the picture \# tell me} & 17 & \colorbox{Green}{mother is \# washing dishes} & 26 \\
 & of \# the & 16 & dishes \# and & 32 & and \# the & 71 & \ovalbox{othing tote \# ofor} & 10 & \colorbox{Green}{mother is \# washing dishes} & 13 & ah it \# looks like & 20 \\
 & dishes \# and & 12 & it \# looks & 27 & of \# the & 60 & ah it \# looks like & 10 & washing dishes \# and the & 12 & the picture \# tell me & 19 \\
 & mother \# and & 11 & i \# see & 25 & the \# little & 54 & it is \# o & 10 & reaching for \# the cookie & 11 & washing dishes \# and the & 18 \\
 & \ovalbox{tote \# ofor} & 10 & \colorbox{Green}{is \# drying} & 19 & i \# see & 50 & \ovalbox{gon as \# aa itre} & 10 & \ovalbox{nitchin cookies \# ok} & 10 & \colorbox{Green}{is open \# looks like} & 17 \\ \hline
\multirow{5}{*}{level 2} & of \# the & 17 & \colorbox{Green}{is \# running} & 38 & and \# the & 117 & or cake \# and look & 10 & mother is \# washing dishes & 24 & \colorbox{Green}{mother is \# washing dishes} & 37 \\
 & it \# looks & 15 & dishes \# and & 38 & the \# sink & 93 & \colorbox{Green}{shink is \# running over} & 10 & \underline{the picture \# well the} & 18 & washing dishes \# and the & 18 \\
 & \colorbox{Green}{is \# drying} & 13 & of \# the & 37 & on \# the & 92 & the mother \# and the & 10 & two cups \# and a & 13 & \underline{the picture \# well the} & 18 \\
 & mother \# and & 12 & and \# the & 32 & of \# the & 76 & \ovalbox{othing tote \# ofor} & 10 & washing dishes \# and the & 12 & of water \# and the & 18 \\
 & \colorbox{Green}{is \# running} & 11 & the \# sink & 27 & i \# see & 66 & ah it \# looks like & 10 & \underline{the picture \# tell me} & 12 & \colorbox{Green}{and the \# sink is } & 18 \\ \hline
\multirow{5}{*}{level 3} & it \# looks & 15 & dishes \# and & 40 & and \# the & 120 & or cake \# and look & 10 & mother is \# washing dishes & 19 & \colorbox{Green}{mother is \# washing dishes} & 37 \\
 & of \# the & 12 & of \# the & 35 & the \# sink & 96 & ah it \# looks like & 10 & \underline{the picture \# well the} & 17 & washing dishes \# and the & 23 \\
 & cake \# and & 10 & \colorbox{Green}{is \# going} & 31 & on \# the & 93 & \ovalbox{little girl \# cubboard doors} & 10 & two cups \# and a & 12 & water on \# the floor & 20 \\
 & dishes \# and & 10 & \colorbox{Green}{is \# running} & 30 & of \# the & 78 & \ovalbox{othing tote \# ofor} & 9 & trying to \# get in & 11 & \underline{the picture \# well the} & 20 \\
 & \ovalbox{girl \# cubboard} & 10 & and \# the & 30 & dishes \# and & 78 & what wit \# the sinks & 9 & \colorbox{Green}{stool is \# going to} & 11 & and the \# sink is & 20 \\ \cline{1-13} 
\end{tabular}%
}
\end{table*}
\section{Evaluation}

In this section, we present the implementation details and report our evaluation results on speech pauses.

\subsection{Implementation details}

We followed the original implementation of WavBERT~\cite{zhu2021wavbert} using PyTorch. The training and the adversarial training used the standard training and testing sets of \ac{ADReSSo}. The training was conducted with 10 rounds, and the average result of 10 rounds was reported. Compared to the original setting, the learning rate was increased to $10^{-5}$, the maximum epoch was reduced to $100$ for faster convergence, and the model accuracy remains similar~\cite{zhu2021wavbert}. 

\subsection{Model accuracy of adversarial attack/training}

Figure~\ref{fig:result} shows the impact on accuracy from adversarial attack and adversarial training.
The original WavBERT used a majority vote of 10 rounds and achieved an accuracy of $83$-$84\%$. The average accuracy of 10 rounds was $82$-$83\%$. We chose to report the average accuracy because it reflects the performance of every individual round. As shown in Figure~\ref{fig:result}(a), with the step-1 attack, the accuracy is lowered to $70$-$72\%$, which shows the WavBERT model may easily be impacted by small perturbation. However, humans in their speeches often add pauses for many unknown reasons, which are not related to their cognitive problems. In addition, we observed that if a multi-step attack adds more perturbation on pauses, the accuracy is significantly lowered. We confirmed that the pause is an important factor for model inference. Lastly, the level-3 model resulted in the lowest accuracy when the step-20 attack was launched. We consider that the level-3 model relies more on pauses to make the inference, and thus is more vulnerable to the adversarial attacks on pauses. Figure~\ref{fig:result}(b) and (c) show the accuracy of models using adversarial training and reversed adversarial training was lowered but not significantly, thus revealing WavBERT's limited understanding of pauses. 

\subsection{Context analysis of pauses}

Table~\ref{tab:word} shows the context analysis results of perturbation. We listed the top five contexts based on the frequency for range $(1,2)$ and attack step $(1, 5, 20)$.  The adversarial samples listed under the column step-$k$ include all the adversarial samples generated from step-$1$ to step-$k$. We have three observations: i) We found the pause between ``is" and a verb is highly dementia-sensitive, as shown in the highlighted green. We confirmed that the long pause between ``is" and a verb only appears in the AD samples but not non-AD samples of the training set. ii) We found interviewers' pause is important, as shown in the underlined text. As this type of pause does not appear in the step-1 attack, it is not considered as the most important pauses for model inference. We consider it still affects the accuracy to some extent. Perez-Toro et al. has reported a similar result~\cite{perez2021influence}. iii) We found ASR misspelling errors are important, such as ``othing tote \# ofor" and ``gon as \# aa itre", as shown in the oval boxes. We consider the misspelling errors introduce the uncommon tokens, which are considered as important context in the downstream task because the pre-trained models have limited knowledge about these tokens.

\begin{table}[h!]
\begin{small}

\centering
\caption{Length analysis. A stands for AD, N for Non-AD}
\label{tab:change}
\renewcommand\arraystretch{1.0}
\resizebox{0.48\textwidth}{!}{%
\begin{tabular}{cccccccc}
\hline
 &  & \multicolumn{2}{c}{Step-1} & \multicolumn{2}{c}{Step-5} & \multicolumn{2}{c}{Step-20} \\
 & Perturbation & A$\rightarrow$N & N$\rightarrow$A & A$\rightarrow$N & N$\rightarrow$A & A$\rightarrow$N & N$\rightarrow$A \\ \hline
\parbox[t]{2mm}{\multirow{2}{*}{\rotatebox[origin=c]{90}{level 1}}} & Add . & 90 & \colorbox{Green}{357} & 495 & \colorbox{Green}{1764} & 1336 & \colorbox{Green}{5715} \\
 & Delete . & \colorbox{Green}{242} & 3 & \colorbox{Green}{844} & 34 & \colorbox{Green}{1168} & 644 \\ \hline
\parbox[t]{2mm}{\multirow{6}{*}{\rotatebox[origin=c]{90}{level 2}}} & Add , & 34 & 1 & 272 & 35 & 1631 & 1365 \\
 & Add . & 38 & \colorbox{Green}{334} & 157 & \colorbox{Green}{1639} & 450 & \colorbox{Green}{4410} \\
 & Replace ,$\rightarrow$. & 5 & 16 & 58 & 86 & 196 & 502 \\
 & Delete , & 57 & 9 & 388 & 19 & 1501 & 369 \\
 & Delete . & \colorbox{Green}{180} & 0 & \colorbox{Green}{591} & 9 & \colorbox{Green}{879} & 188 \\
 & Replace .$\rightarrow$, & 36 & 0 & 284 & 12 & 653 & 243 \\ \hline
\parbox[t]{2mm}{\multirow{12}{*}{\rotatebox[origin=c]{90}{level 3}}} & Add , & 36 & 2 & 284 & 15 & 1686 & 629 \\
 & Add ; & 14 & 21 & 58 & 123 & 303 & 1009 \\
 & Add . & 34 & \colorbox{Green}{311} & 111 & \colorbox{Green}{1524} & 329 & \colorbox{Green}{4118} \\
 & Replace ,$\rightarrow$; & 0 & 5 & 4 & 9 & 89 & 141 \\
 & Replace ,$\rightarrow$. & 2 & 10 & 19 & 89 & 86 & 454 \\
 & Replace ;$\rightarrow$. & 3 & 2 & 29 & 9 & 72 & 88 \\
 & Delete , & 8 & 7 & 136 & 13 & 946 & 218 \\
 & Delete ; & 41 & 0 & 187 & 0 & 444 & 75 \\
 & Delete . & \colorbox{Green}{153} & 2 & \colorbox{Green}{488} & 11 & \colorbox{Green}{719} & 163 \\
 & Replace ;$\rightarrow$, & 9 & 0 & 92 & 1 & 467 & 64 \\
 & Replace .$\rightarrow$, & 42 & 0 & 284 & 5 & 630 & 82 \\
 & Replace .$\rightarrow$; & 8 & 0 & 58 & 1 & 169 & 73 \\ \hline
\end{tabular}%
}
\end{small}
\end{table}

\subsection{Length analysis of pauses}

Table~\ref{tab:change} shows the frequency of perturbation actions. Adding and deleting the long pauses (punctuation ``period") are common perturbation methods to change detection results (highlighted). In step-1 attack, we consider the perturbation actions are performed on sensitive pauses. In cases of changing confidence towards non-AD, the total number of deleting and long-to-short replacing actions is significantly larger than the total number of adding and short-to-long replacing. In cases of changing confidence towards AD, the perturbation actions are mostly adding pauses. However, we note that most actions of adding a comma in level 2/3 and adding a period in level 1 lead to a confidence change towards non-AD. Thus, we consider that not all pauses are dementia-sensitive from the model's perspective, because natural language contains pauses. Lastly, we found that replacing actions appear less frequently than adding and deleting, suggesting that the pause context is more important than the pause length.

\section{Conclusions}
In this paper, we exploited adversarial learning to study the interpretability of speech pauses in dementia detection. With the adversarial attack and training techniques, we successfully identified the positions and lengths of sensitive pauses, e.g., speech pauses near to the verb “is” are dementia-sensitive. We also confirmed that increasing lengths of sensitive pauses or adding sensitive pauses leads the model inference to AD, while decreasing the lengths of sensitive pauses or deleting sensitive pauses leads to non-AD. Our evaluation reveals that the misspelling errors are inevitable if using speech recognition and language model. One future direction is to exploit large-scale speech data for pre-training a speech-based language model, thus resolving the misspelling problem.

\bibliographystyle{IEEEbib}
\bibliography{main}

\section{List of Acronyms}
\begin{acronym}
\acro{ADRD}{Alzheimer's Disease and Related Dementias}
\acro{AD}{Alzheimer's Disease}
\acro{MCI}{Mild Cognitive Impairment}
\acro{HC}{Healthy Control}
\acro{WLS}{Wisconsin Longitudinal Study}
\acro{CTP}{Cookie Theft Picture}
\acro{IVA}{Intelligent Virtual Agent}
\acro{IU}{Information Units}
\acro{MFCC}{Mel Frequency Cepstral Coefficient}
\acro{LLDs}{Low-Level Descriptors}
\acro{LSP}{Line Spectral Pair}
\acro{AOI}{Area of Interest}
\acro{ASR}{Automatic Speech Recognition}
\acro{ML}{Machine-Learning}
\acro{MMSE}{Mini-Mental State Examination}
\acro{MoCA}{Montreal Cognitive Assessment}
\acro{GDS}{Geriatric Depression Scale}
\acro{GAI}{Geriatric Anxiety Inventory}
\acro{SVM}{Support Vector Machine}
\acro{PCA}{Principal Component Analysis}
\acro{DNN}{Deep Neural Network}
\acro{MECSD}{Mandarin Elderly Cognitive Speech Database}
\acro{LM}{Language Model}
\acro{DNN}{Deep Neural Network}
\acro{FCN}{Fully Convolutional Network}
\acro{CNN}{Convolutional Neural Network}
\acro{GAP}{Global Average Pooling}
\acro{FC}{Fully Connected}
\acro{OARS}{Older Americans Resources and Services}
\acro{LDA}{Latent Dirichlet Allocation}
\acro{ADReSS}{Alzheimer's Dementia Recognition through Spontaneous Speech}
\acro{ADReSSo}{ADReSS speech only}
\acro{SVF}{Semantic Verbal Fluency}
\acro{VAS}{Voice-Assistant System}
\acro{RA}{Research Assistant}
\acro{AI}{Artificial Intelligence}
\acro{WER}{Word Error Rate}
\acro{MER}{Match Error Rate}
\acro{1NN}{1-hidden-layer Neural Network}
\acro{2NN}{2-hidden-layer Neural Network}
\acro{DT}{Decision Tree}
\acro{RF}{Random Forest}
\acro{RMSE}{Root-Mean-Square Error}
\acro{LOSO}{Leave-One-Subject-Out}
\acro{IRB}{Institutional Review Board}
\acro{BERT}{Bidirectional Encoder Representations from Transformers}
\acro{CTC}{Connectionist Temporal Classification}
\acro{MLM}{Masked Language Model}
\acro{NSP}{Next Sentence Prediction}
\acro{NLP}{Natural language processing}
\acro{MRI}{Magnetic Resonance Imaging}
\end{acronym}
\end{document}